\documentclass[runningheads]{llncs}
\usepackage[T1]{fontenc}
\usepackage{graphicx}
\usepackage{array}
\usepackage{xcolor}
\usepackage{subcaption}
\usepackage{amsmath}
\usepackage{float}
\usepackage{placeins}
\usepackage{multirow}
\usepackage{subfiles}
\usepackage{hyperref}

\usepackage[capitalize,nameinlink]{cleveref}

\begin{document}
\title{FIORD: A Fisheye Indoor-Outdoor Dataset with LIDAR Ground Truth for 3D Scene Reconstruction and Benchmarking}
\titlerunning{FIORD}
%
\authorrunning{U. Gunes et al.}
\author{
  Ulas Gunes\inst{1} \and
  Matias Turkulainen\inst{2} \and
  Xuqian Ren\inst{1}\and
  Arno Solin\inst{2} \and
  Juho Kannala\inst{2,3} \and
  Esa Rahtu\inst{1}
}

\institute{
  Tampere University, Finland \\ \email{\{ulas.gunes, xuqian.ren, esa.rahtu\}@tuni.fi}
  \and
  Aalto University, Finland \\
  \email{\{matias.turkulainen, arno.solin, juho.kannala\}@aalto.fi}
    \and
  University of Oulu, Finland
}
\maketitle              
\begin{abstract}
The development of large-scale 3D scene reconstruction and novel view synthesis methods mostly rely on datasets comprising perspective images with narrow fields of view (FoV). While effective for small-scale scenes, these datasets require large image sets and extensive structure-from-motion (SfM) processing, limiting scalability. To address this, we introduce a fisheye image dataset tailored for scene reconstruction tasks. Using dual 200-degree fisheye lenses, our dataset provides full 360-degree coverage of 5 indoor and 5 outdoor scenes. Each scene has sparse SfM point clouds and precise LIDAR-derived dense point clouds that can be used as geometric ground-truth, enabling robust benchmarking under challenging conditions such as occlusions and reflections. While the baseline experiments focus on vanilla Gaussian Splatting and NeRF based Nerfacto methods, the dataset supports diverse approaches for scene reconstruction, novel view synthesis, and image-based rendering. The dataset is available \href{https://zenodo.org/communities/fiord}{here}.

\keywords{fisheye image dataset \and 3D scene reconstruction \and novel view synthesis \and Gaussian splatting \and image-based rendering}
\end{abstract}
\begin{figure}[ht]
  \centering
  \includegraphics[width=0.9\textwidth]{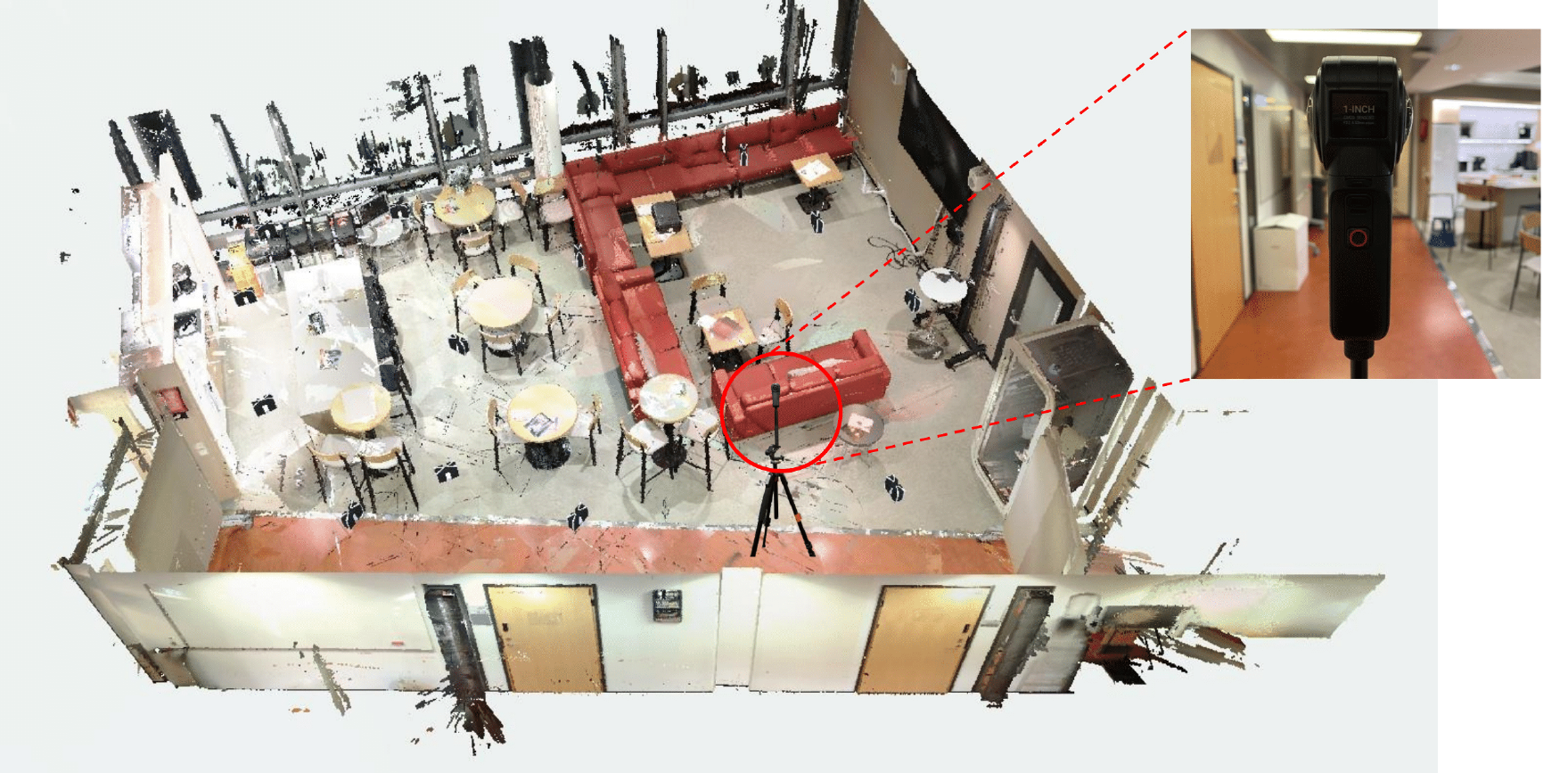}
  \vspace{0.5em}
  \includegraphics[width=0.9\textwidth]{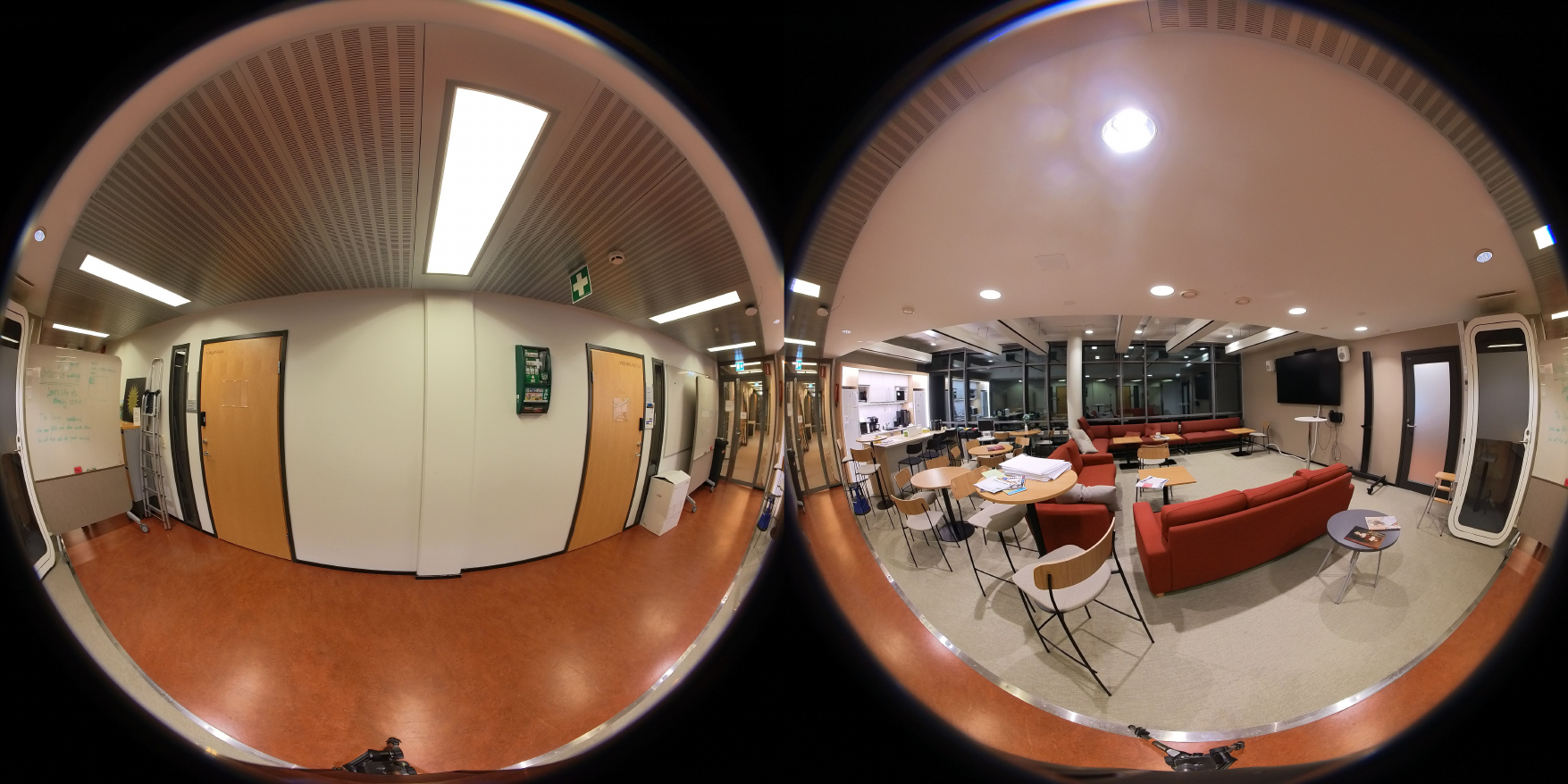}
  \caption{\textbf{Example data capture setup.} The top image shows an example placement of camera in one of our scenes (depicted as its dense point cloud from Faro scanner), while the bottom image illustrates wide-angle 360\textdegree\ photo captured with a single shot of the camera.}
  \label{fig:1}
\end{figure}
\vspace{-1cm}
\section{Introduction}
Recent advances in computer vision and graphics have revolutionized 3D scene reconstruction, novel view synthesis, and image-based rendering. Techniques such as 3D Gaussian splatting (3DGS, \cite{GS}), with its explicit point-based representation, and Neural Radiance Fields (NeRFs, \cite{mildenhall2020nerf}) have demonstrated remarkable results in these tasks. Gaussian splatting, in particular, offers faster rendering speeds and scalability for real-time applications. Unlike NeRFs, which encode volumetric data in neural networks, Gaussian splatting directly models scene geometry and appearance using Gaussian primitives, making it highly efficient for large-scale reconstruction.

Existing datasets commonly used for developing these techniques are predominantly composed of perspective images with narrow fields of view (FoV, typically <120\textdegree) and they are tailored for specific applications such as object-centric scenes or urban driving scenarios. Additionally, these datasets are often captured as videos during movement in the environment, introducing motion blur and compromising image quality. These limitations make them less suitable for broader applications in 3D reconstruction, novel view synthesis, and image-based rendering, particularly in large-scale, non-object-centric environments.

To address these challenges, we introduce a high-resolution, ultra-wide-angle fisheye dataset designed to support a wide range of applications. Additionally, we include precise LIDAR-derived ground truth point clouds of the captured environments using a Faro Focus 3D scanner \cite{faro_focus}. Baseline evaluation results for vanilla Gaussian splatting and Nerfacto on this dataset are also provided, serving as a reference for future methods.

Our contributions are as follows:
\begin{itemize}
    \item \textbf{A high-resolution, wide-angle fisheye still image collection:} The wide-angle images provide comprehensive scene coverage with fewer captures compared to narrow-field-of-view perspective images. The use of static photographs avoids motion blur associated with video frame extraction, ensuring sharp feature matching and improved reconstruction accuracy in Structure-from-Motion (SfM) pipelines.

    \item \textbf{Dense LIDAR-derived ground truth:} Dense point clouds generated with a Faro Focus 3D laser scanner serve as authoritative references for evaluating and improving reconstruction pipelines, particularly for alignment-sensitive techniques like Gaussian Splatting and NeRF.

    \item \textbf{SfM-compatible sparse point clouds:} Sparse point clouds generated with the Structure-from-Motion (SfM) tool COLMAP \cite{schoenberger2016sfm,schoenberger2016mvs,schoenberger2016vote} are included in the dataset.

    \item \textbf{Baseline evaluations and benchmarks:} We provide baseline results for vanilla Gaussian splatting and Nerfacto methods, which offer insight into the potential of the dataset for novel view synthesis and scene reconstruction tasks. These benchmarks can guide future research and serve as references for evaluating new rendering, reconstruction, and depth-based methods.
\end{itemize}
By addressing the limitations of existing datasets, our work enables the development of novel techniques for diverse real-world scenarios in 3D reconstruction, image-based rendering, and novel view synthesis.

\section{Related Work}
Out of the numerous datasets available for 3D reconstruction and novel view synthesis tasks, we present a curated selection here, chosen for their diversity, scene coverage complexity and modality, while noting that many can also be repurposed for broader computer vision applications.
\paragraph{Datasets for 3D Scene Reconstruction and Novel View Synthesis.} 
The Tanks and Temples \cite{Knapitsch2017} dataset provides high-quality ground truth data derived from an industrial laser scanner and high-resolution video input for both indoor and outdoor settings, and serves as a benchmark for static scene reconstruction. Waymo Open Dataset \cite{Sun_2020_CVPR}, KITTI-360 \cite{Liao2022PAMI}, and nuScenes \cite{nuscenes} similarly provide extensive multi-modal sensor data, including LIDAR, RGB imagery, and trajectory information, making them suitable for scene reconstruction tasks. 

The ScanNet++ \cite{yeshwanthliu2023scannetpp} dataset extends the original ScanNet \cite{dai2017scannet} dataset by adding object-level semantics and refining camera pose alignments, making it particularly suitable for semantic and geometric indoor reconstructions. MuSHRoom~\cite{ren2023mushroom} emphasizes diverse indoor environments, captured using high-precision and consumer-grade sensors. Replica~\cite{replica19arxiv} offers photorealistic reconstructions of indoor spaces, widely used in visual SLAM and neural rendering. The dataset used in Mip-NeRF360 \cite{mipnerf360} is another well-known one, which focuses on unbounded, object-centric scenes and small-scale, bounded indoor environments, accompanied by estimated camera poses from COLMAP.

The Aerial Coastline Imagery Dataset (ACID, \cite{infinite_nature_2020}) captures natural coastal scenes using aerial drone footage, which allows long-range trajectory synthesis in scenes. UrbanScene3D \cite{UrbanScene3D} similarly facilitates bird-eye view urban reconstructions and was pivotal in an early large-scale 3DGS-based work \cite{lin2024vastgaussianvast3dgaussians}. MatrixCity \cite{li2023matrixcity} also delivers synthetic data for controlled Gaussian splatting experiments across ground-level and aerial scenes, which have been used in another early work on the 3DGS-based large-scale scene reconstruction \cite{liu2024citygaussian}. Similar to our work, the dataset used in the hierarchical Gaussian splatting \cite{hierarchicalgaussians24}, collected with a multi-camera GoPro rig that captures time-elapsed narrow FoV images in motion (walking on foot or moving by bicycle), focuses on large-scale scene modeling.

The 360Roam dataset \cite{huang2022360roam} provides full 360\textdegree imagery optimized for Gaussian splatting, while EgoNeRF \cite{egonerf} dataset focuses on omnidirectional modeling for large-scale indoor reconstructions. OmniGS~\cite{omnigs} work leverages the panoramic datasets 360Roam~\cite{huang2022360roam} and EgoNeRF~\cite{egonerf} for indoor reconstructions, highlighting the utility of omnidirectional data for large-scale modeling in panoramic format.  The LetsGo project \cite{letsgo} uses a commercial LIDAR and fisheye imaging device in the same coordinate system for garage-scale environments, which have bounded (indoor) and semi-bounded (garage with outdoor openings) scene components. Unlike these works, our dataset captures images in raw fisheye format rather than the equirectangular format, which is commonly used to stitch the two fisheye views from a 360\textdegree camera. This approach eliminates potential stitching artifacts that may arise during the transformation process. Additionally, our dataset is captured in a completely motion-free setting, ensuring that each frame remains still and unaffected by movement, in contrast to datasets that rely on moving systems that record videos or capture time-elapsed images.
\paragraph{Fisheye Image Based Rendering.} Recent works addressing the challenges of rendering wide-angle fisheye images (>180\textdegree FoV) include On the Error Analysis of 3D Gaussian Splatting \cite{errorgs}, which introduces a rasterizer for fisheye rendering without rectification, and 3DGUT \cite{wu20243dgutenablingdistortedcameras}, which extends 3D Gaussian splatting to support non-linear camera projections and secondary rays for simulating effects like reflections and refractions. Both methods demonstrate these capabilities using indoor and unbounded fisheye images.
\vspace{-2mm}
\section{The FIORD Dataset}
\vspace{-2mm}
The FIORD comprises of still fisheye images captured from ten distinct scenes, provided in their stitched (two fisheyes side to side) and split (single fisheye) formats. Additionally, the dataset includes two types of point clouds for each scene: the sparse point cloud generated via Structure-from-Motion (SfM) and the dense point cloud captured with a Faro LIDAR scanner.

\begin{figure}[t]
  \centering
  \includegraphics[width=\textwidth]{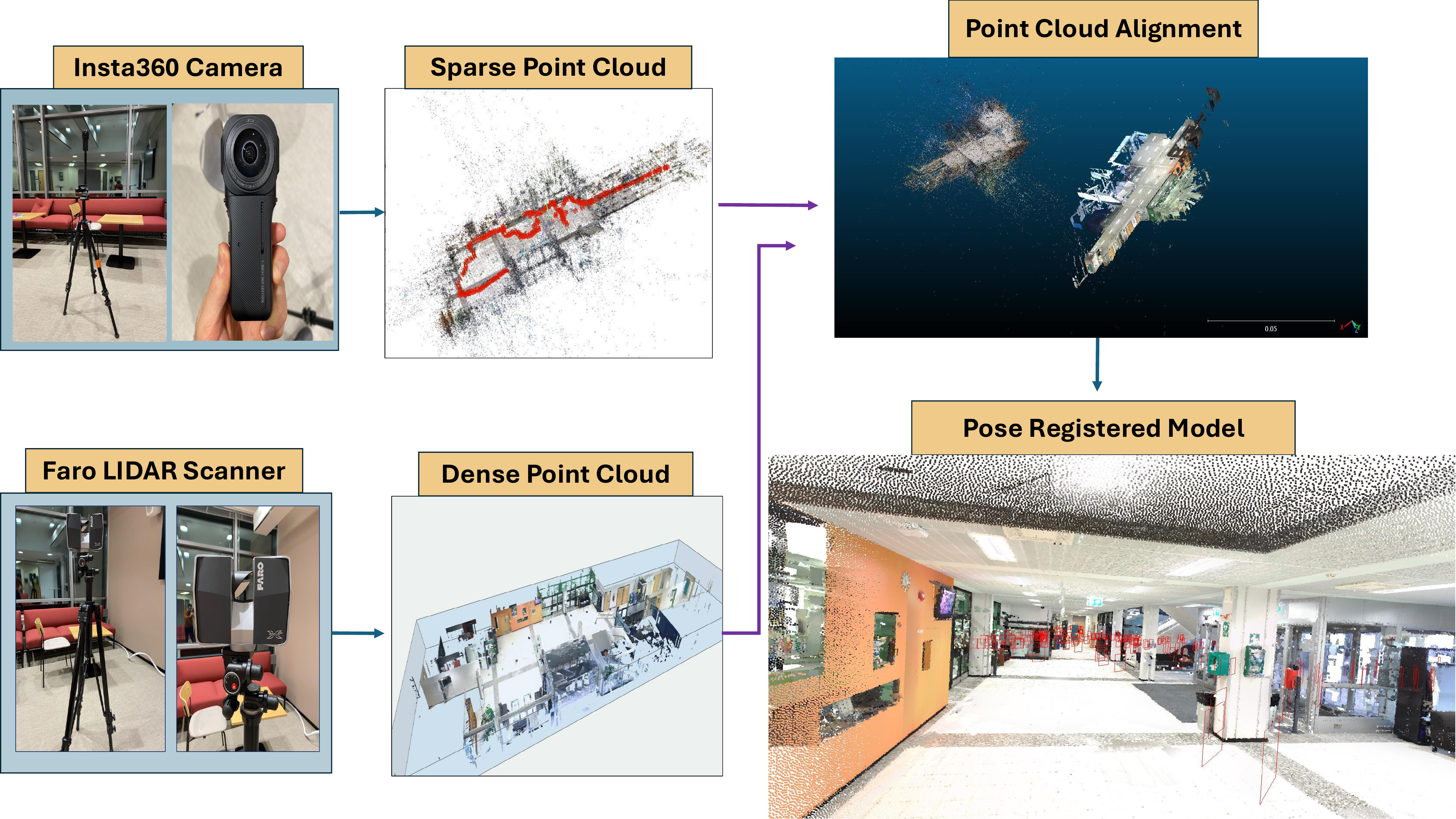} 
  \caption{\textbf{Sparse and dense point cloud alignment}. Fisheye images and LIDAR scans are used to generate and align sparse and dense point clouds. Camera poses can be registered to the aligned model in real-world or COLMAP coordinate system scale.}
  \label{fig:2}
\end{figure}

In this section, we first explain the camera calibration and data collection procedures performed using the Insta360 RS One-Inch camera \cite{insta360} and the Faro Focus LIDAR Scanner. Then, we explain the post-data collection processing steps to create our dataset, which yields the generation of sparse and dense point clouds for the scenes. Finally, we describe the sparse and dense point cloud alignment process and the image rectification step for our experiments mentioned in the next section. The overall process is visualized in \cref{fig:2}.

\subsection{Camera Calibration}

To prepare this data set, we used two wide-angle (200\textdegree FoV per lens) fisheye cameras of the Insta360 One RS One-Inch Sensor camera \cite{insta360}, mounted on a tripod. In other words, all the images we captured, both for scene data collection and calibration purposes, are still (motion-free). Although the Insta360 camera can stitch the images from its two fisheye lenses to generate full 360\textdegree equirectangular panoramic images, this stitching process often introduces alignment errors, such as ghosting artifacts along stitching lines, particularly in scenes with complex geometry or significant depth variation. To avoid these issues and preserve geometric accuracy, we opted to work directly with the raw fisheye images.

The raw fisheye images were initially stored in the camera manufacturers' .insp file format, and we converted them to JPEG. JPEG was chosen because it is supported by both the Camera Calibration Toolbox and SfM software, and it offers a practical balance of compatibility, quality, and file size. Higher-fidelity formats like PNG could be used for tasks requiring lossless quality, albeit with increased computational cost. Each capture results in a single image that contains two side-by-side fisheye views (\cref{fig:1}). These images were split into two separate 3264$\times$3264 images, each corresponding to one fisheye lens.

 Fisheye lenses produce heavily distorted images, particularly toward the edges, where the distortion effect becomes most visible (\cref{fig:1}). We performed camera calibration to correct these distortions. We extracted precisely estimated intrinsic camera parameters for each fisheye lens separately.

Using the estimated intrinsic camera parameters, we performed camera calibration. We utilized the Camera Calibration Toolbox for Generic Lenses \cite{KannalaBrandt}\cite{KannalaHB08}, which provides calibration support for wide-angle lenses exceeding 180\textdegree FoV. We used the radial camera model option from the toolbox for our lenses. The calibration pattern displayed on a flat screen was captured from both the heavily distorted edges and the less distorted central region of each lens by placing the tripod-mounted camera in different locations in front of the pattern. Camera exposure, shutter speed, and white-balancing settings were kept constant while capturing still images of the calibration target.

After the calibration process, we converted the intrinsic camera parameters we obtained from the toolbox to OpenCV-Fisheye camera format \cite{opencv_fisheye}. This format includes focal lengths $(f_x,f_y)$, principal point coordinates $(c_x,c_y)$, and radial distortion coefficients $(k_1,...,k_4)$, and is accepted by the SfM pipeline COLMAP \cite{schoenberger2016sfm,schoenberger2016mvs}, which will be used in the next steps. We verified the accuracy of the calibration by using the estimated intrinsic parameters to remove the radial distortion from our fisheye images.

\subsection{Data Collection}

To create a diverse dataset, we selected ten distinct scenes (five indoor environments and five outdoor environments) from the Tampere University Hervanta Campus, with varying color, lighting characteristics, scale, and geometric complexity. Depending on the scene size and complexity, a total of 500--1300 images (single fisheye) per scene are captured. Our scenes were captured during winter conditions, which introduced unique realistic challenges for our outdoor environments, such as snow or ice-induced glare, foggy conditions, occlusions, complex lighting conditions and repetitive structures that might be challenging for the SfM pipeline based sparse point cloud reconstruction. We believe that these challenges will push forward the research in 3D scene reconstruction techniques. The scenes in our dataset and their brief descriptions are given in \cref{newtable}. 

\begin{table}[ht]
\centering
\caption{Descriptions of Indoor and Outdoor Scenes in the Dataset}
\label{newtable}
\begin{tabular}{|l|p{0.75\textwidth}|}
\hline
\multicolumn{2}{|c|}{\textbf{Indoor Scenes}} \\ \hline
\textbf{Name}          & \textbf{Description} \\ \hline
Kitchen\_In                & A 12m\textsuperscript{2} kitchen featuring repetitive, detailed objects (chairs), appliances, and reflective countertops, providing moderate geometric complexity. \\ \hline
MeetingRoom\_In           & A 15 m\textsuperscript{2} room with simple geometry, flat walls, and minimal objects, with heavy ceiling light exposure. \\ \hline
Building\_In         & A hallway-like 62m\textsuperscript{2} indoor environment with uniform light distribution, repetitive tiles and reflective materials. \\ \hline
Hall\_In        & A large 80m\textsuperscript{2} hallway with tall ceiling, nonuniform light distribution, repetitive tiles, shiny and or highly detailed objects. \\ \hline
Upstairs\_In      & A large 260 m\textsuperscript{2} hall area with irregular shapes, textured surfaces, and reflective materials such as glass. \\ \hline
\multicolumn{2}{|c|}{\textbf{Outdoor Scenes}} \\ \hline
\textbf{Name}          & \textbf{Description} \\ \hline
Bridge\_Out          & A 125m\textsuperscript{2} outdoor walkway with snow, reflective glasses and repetitive texture buildings. \\ \hline
Night\_Out          & A 125m\textsuperscript{2} outdoor garden area with trees, buildings with reflective surfaces, repetitive window patterns and non-uniform light. Captured during evening conditions. \\ \hline
Corridor\_Out            & A 207m\textsuperscript{2} outdoor walkway with snow, repeptitive structured stairs, glasses and non-homogenous (pepper-salt style small rocks) floor structure. \\ \hline
Building\_Out              & A 305m\textsuperscript{2} outdoor space, includes a couple of moving objects such as people or cars. \\ \hline
Road\_Out              & A 930m\textsuperscript{2} large unbounded outdoor space with occlusions, non-uniform light conditions, non-homogenous (pepper-salt style small rocks) floor structure, reflective surfaces and fog. \\ \hline
\end{tabular}
\end{table}

The Insta360 camera \cite{insta360} was mounted on a tripod, and placed at a fixed location to take a single shot for the scene image capturing process (an example camera placement in a scene is shown in \cref{fig:1}). Afterwards, it was repositioned with minimal rotation and movement (less than 10 cm between each camera positioning and less than 60 degrees of rotation to either side on the lateral axis of the camera) to another location within the scene, and another image was taken. This process was repeated systematically until the entire scene was covered, with each lens consistently covering the same side of the scene at all times. This consistency, combined with the minimal rotation or movement between each capture ensured sufficient overlap between the images, which played a key role in the subsequent accurate SfM based sparse point cloud generation step. The short focal length of the camera also enabled sharp capture of wide areas, even from long distances. Similar to the calibration step, consistent settings for shutter speed, exposure, and white balance during image captures are used to capture true lighting and color in scenes.

To avoid disruptions from moving objects, such as people in indoor environments or cars in outdoor settings, images were captured at times and locations with minimal activity. The photographer ensured they remained out of the frame by strategically positioning themselves in occluded areas or sequentially capturing the two fisheye images from the same fixed position—first taking a shot while remaining outside the field of view (FoV) of one lens, then repositioning to avoid the FoV of the second lens before capturing the next image.

To obtain the geometry ground truth of each scene, we used the Faro Focus 3D LIDAR scanner fixed on a tripod, to capture high-resolution XYZRGB point clouds. The Faro scanner covers a 360\textdegree horizontal and 170\textdegree vertical FoV ($-60$\textdegree to 90\textdegree) with a data capture range of 0.6–200 meters depending on the indoor or outdoor capture modes. Similar to the camera setup, the Faro scanner was placed at fixed locations in the scene and multiple scans, each corresponding to a different location in the scene, are taken at 1/4 or 1/5 resolution and $4\times$ quality. Each scan lasted around 11 minutes and depending on the scale of the scene, 5--25 scans at different fixed positions were necessary to fully cover each scene. Minor artifacts from moving objects were negligible relative to scene scale and were not visible in the dense point clouds created. An example dense point cloud captured with the Faro Scanner for the Kitchen\_In scene is given in Figure \ref{fig:1}. 

\subsection{Formation of Sparse and Dense Point Clouds}
The Structure-from-Motion (SfM) is a fundamental step in many 3D scene reconstruction methods, as it allows for the recovery of camera poses and the forming of sparse 3D structures from multiple overlapping images. In our dataset, we utilized COLMAP \cite{schoenberger2016sfm}\cite{schoenberger2016mvs} version 3.9.1, an incremental SfM pipeline, to generate sparse point clouds of the scenes using the raw fisheye images.

As the first step of SfM, feature extraction, in COLMAP, we employed the OpenCV Fisheye camera model and supplied the original calibration parameters. To accommodate the high-resolution fisheye images, we increased the maximum image size allowed and the number of features extracted. After extracting feature points from the images, feature matching was performed using the vocabulary tree matcher. In COLMAP, images are registered into the scene representation, followed by triangulation of 3D points and a global bundle adjustment to simultaneously optimize camera poses and the 3D structure. This reconstruction process results in a sparse point cloud.

The COLMAP SfM pipeline outputs the sparse point cloud representing the 3D structure of each scene in binary (.bin) format. The binary format is included in each scenes' sparse model in our dataset, and these can be converted to the text format if necessary. The COLMAP format is also supported by Nerfstudio \cite{nerfstudio}, a common 3D scene reconstruction framework that can enable real-time rendering in navigable environments for a number of NeRF and 3DGS-based scene reconstruction methods. 

The dense, ground-truth point cloud is obtained from the Faro scanner software SCENE \cite{faro_focus}, after the software processes the scans we have taken for each scene. Minimal cropping operations are performed to remove redundant points. 

\subsection{Point Cloud Alignment and COLMAP Model Modification}
Aligning sparse point clouds generated by COLMAP with dense ground truth data from the Faro scanner establishes a shared coordinate system and enables direct comparison and evaluation for downstream applications. 

We performed an alignment between the two point clouds, using the CloudCompare \cite{cloudcompare2024} software. First we selected 7–10 easily identifiable points (e.g., corners, edges, and structural features) from the COLMAP point cloud and then have marked their correspondences in the FARO Scanner generated point cloud. Based on these correspondences, we estimated a transformation matrix defining the rotation, translation, and scaling required to map the sparse COLMAP point cloud to the Faro scan’s coordinate system. 

A key challenge in this process is the significant difference in the density of the point clouds. For example, in the largest indoor scene, the sparse point cloud produced by COLMAP contains about 400,000 points, while the dense Faro scan for the same scene includes nearly 500 million points. This disparity complicates the task of mapping corresponding points between the two datasets. 

The alignment accuracy was validated using the Root Mean Square Error (RMSE) metric calculated in CloudCompare. RMSE quantifies the mean distance between these corresponding points in the two ppoint clouds, indicating how closely the two clouds overlap after alignment. For instance, an RMSE of 25 cm in our largest indoor scene suggests that, on average, the corresponding points in the sparse and dense clouds differ by 25 cm, which is acceptable for scenes of this scale (e.g., a 220m\textsuperscript{2} room). In addition to the RMSE metric, the alignment was verified visually. The final transformation matrix was applied to the COLMAP model, updating all reconstructed points and camera poses. A simplistic illustration of the alignment process is given in \cref{fig:2}.
\vspace{-2mm}
\subsection{Image Rectification}
\vspace{-2mm}
For compatibility with our experiments presented in the next section, we performed image rectification (undistortion) on the raw fisheye images using the estimated intrinsic camera calibration parameters we obtained in Section 3.1. This step transformed the fisheye images into a pinhole camera model-compatible format, and facilitated our baseline experiments using Gaussian Splatting \cite{GS} and Nerfstudio's Nerfacto \cite{nerfstudio} methods, which lack native support for fisheye rendering for wide-angle lenses (>180\textdegree FoV). 

The rectification process was implemented for convenience and allowed the dataset to be easily integrated into the novel 3D scene reconstruction pipelines. However, this approach is not optimal, as the rectified images may lose scene information from the heavily distorted parts of the fisheye images. Despite this, the processed dataset provides a practical solution for our experiments and works sufficiently well, based on the quantitative results.
\vspace{-3mm}
\section{Experiments} 
\vspace{-2mm}
In this section, we present two experiments that demonstrate the capabilities of our dataset, highlighting the usage of both the sparse COLMAP data and the dense Faro scanner data. For our experiments, we applied a standard 90\%-10\% train-test split to the images for each scene. The visual results and evaluation metrics presented correspond to the rendered test images, which were randomly selected from the complete set of images for each scene.

\subsection{Novel View Synthesis with Vanilla Gaussian Splatting (3DGS) and Nerfacto Using SfM (Sparse) Point Cloud}

In the first experiment, we establish a baseline for rendering quality and performance using the Gaussian splatting (3DGS) \cite{GS} and Nerfstudio (v1.1.4)'s NeRF-based scene reconstruction method Nerfacto \cite{nerfstudio}. For both of these models we use the sparse point clouds generated by the SfM software COLMAP. These point clouds originate from fisheye images, which we rectify (undistort) using camera calibration parameters to be compatible with the Gaussian splatting implementation’s rasterization requirements for pinhole camera models \cite{GS}, and with the Nerfacto models' supported camera types.

For training the Gaussian splatting and Nerfacto models, we down-sample the high-resolution images by 4 ($800\times800$ pixels per image), because processing them at full resolution exceeds available RAM capacity. All tunable parameters for these models remained at their default values. Training and evaluation were carried out on an NVIDIA RTX 4090 GPU, which has 24~GB VRAM. The highest VRAM consumption was recorded as 13~GB for training the Nerfacto model for our largest outdoor scene, Road\_Out. The training and evaluation pipeline took between 20 to 30 minutes for Gaussian splatting model, and between 7--15~minutes for the Nerfacto model depending on scene complexity. We trained each model for 30k iterations for each scene.

\begin{figure}[t]
\centering
\small
\makebox[\textwidth][l]{\hspace{-7mm}
\begin{minipage}{\textwidth}
    \begin{minipage}{0.1\textwidth} 
        \raggedleft
        \vspace{6mm} 
        \rotatebox{90}{\textbf{Ground Truth}} \\[6mm]
        \vspace{-1mm} 
        \rotatebox{90}{\textbf{3DGS Renders}}
    \end{minipage}
    \hfill
    \begin{minipage}{0.88\textwidth} 
        \centering
        \begin{minipage}{0.24\textwidth}
            \centering
            \textbf{Bridge\_Out} \\[1mm]
            \includegraphics[width=\textwidth]{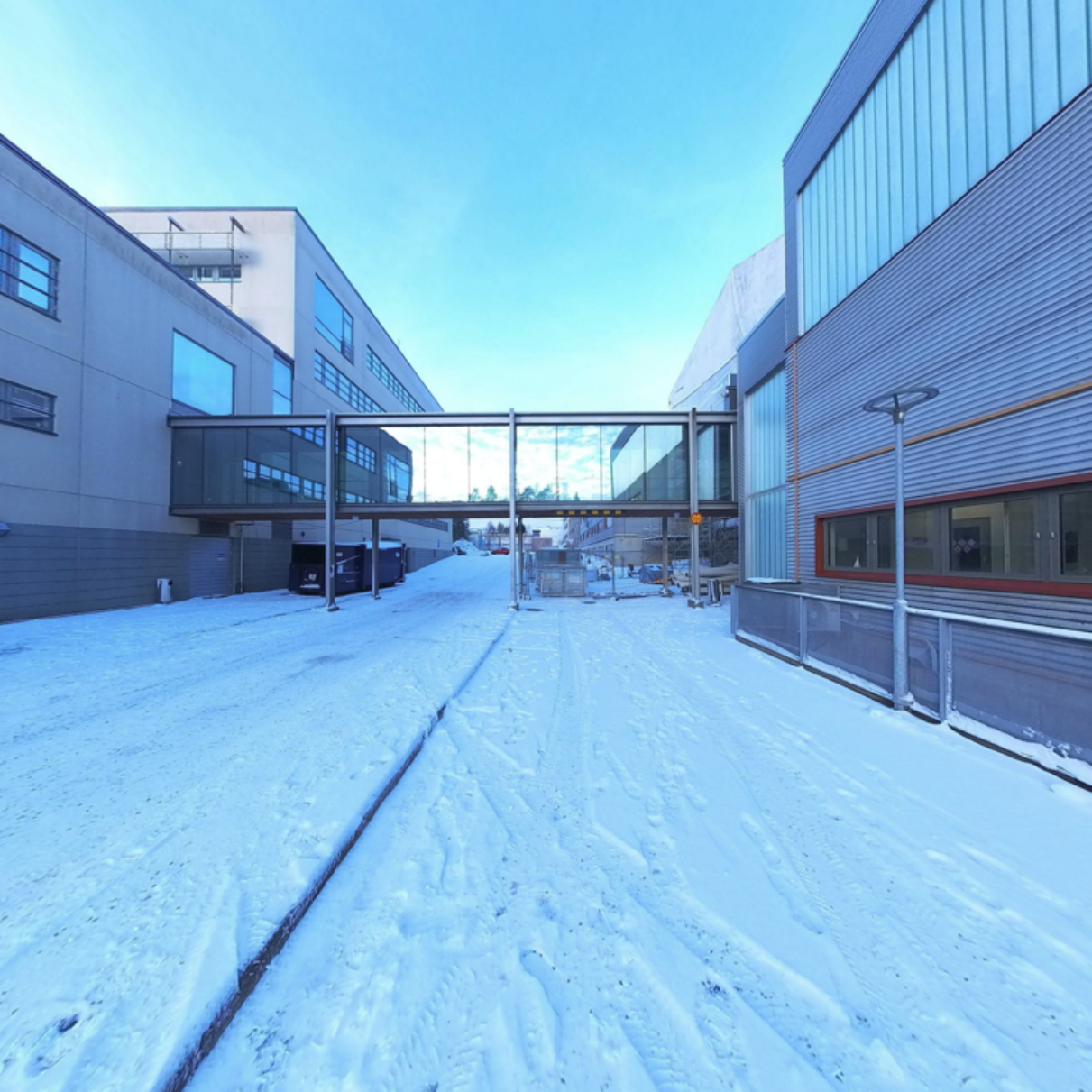}
        \end{minipage}
        \hfill
        \begin{minipage}{0.24\textwidth}
            \centering
            \textbf{Corridor\_Out} \\[1mm]
            \includegraphics[width=\textwidth]{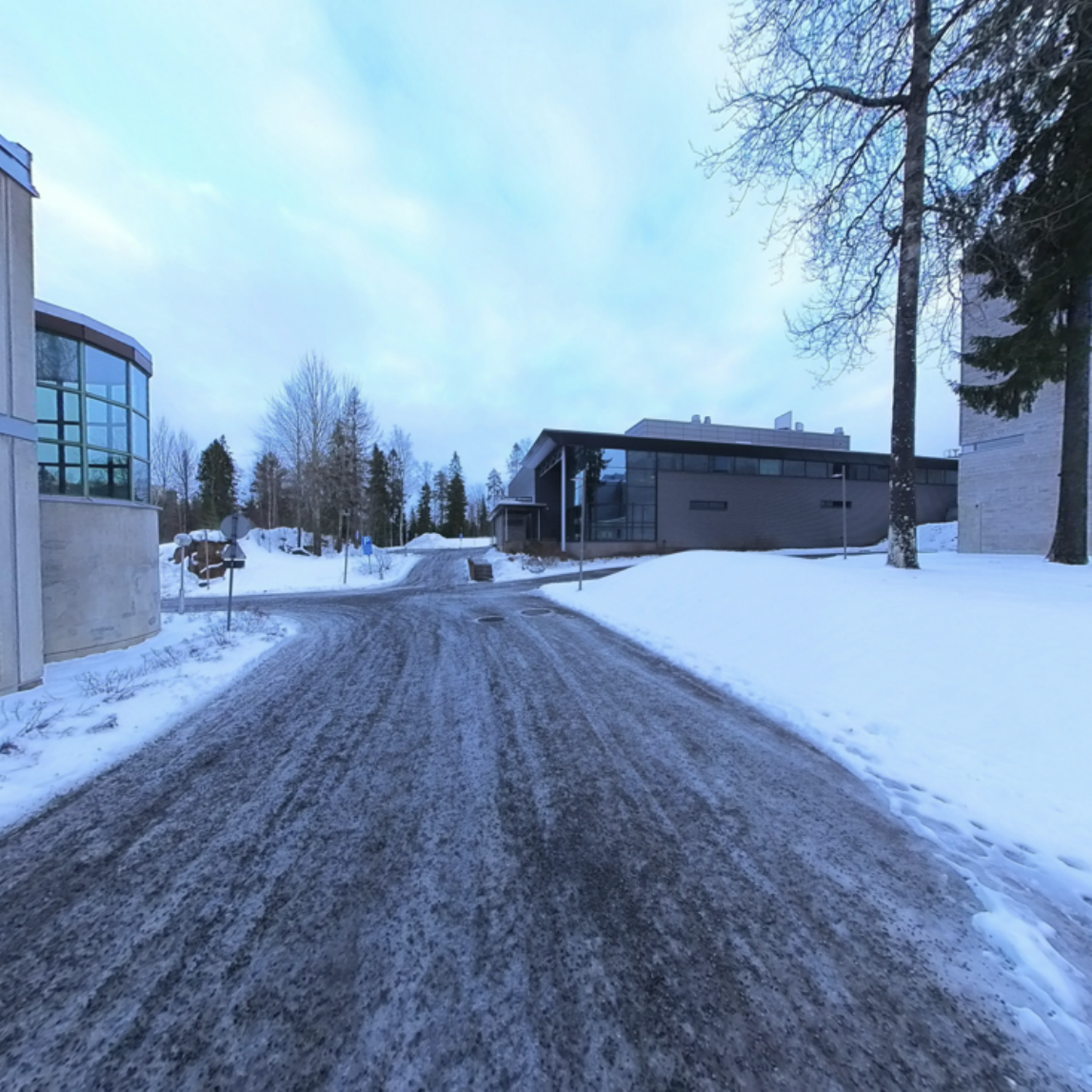}
        \end{minipage}
        \hfill
        \begin{minipage}{0.24\textwidth}
            \centering
            \textbf{Upstairs\_In} \\[1mm]
            \includegraphics[width=\textwidth]{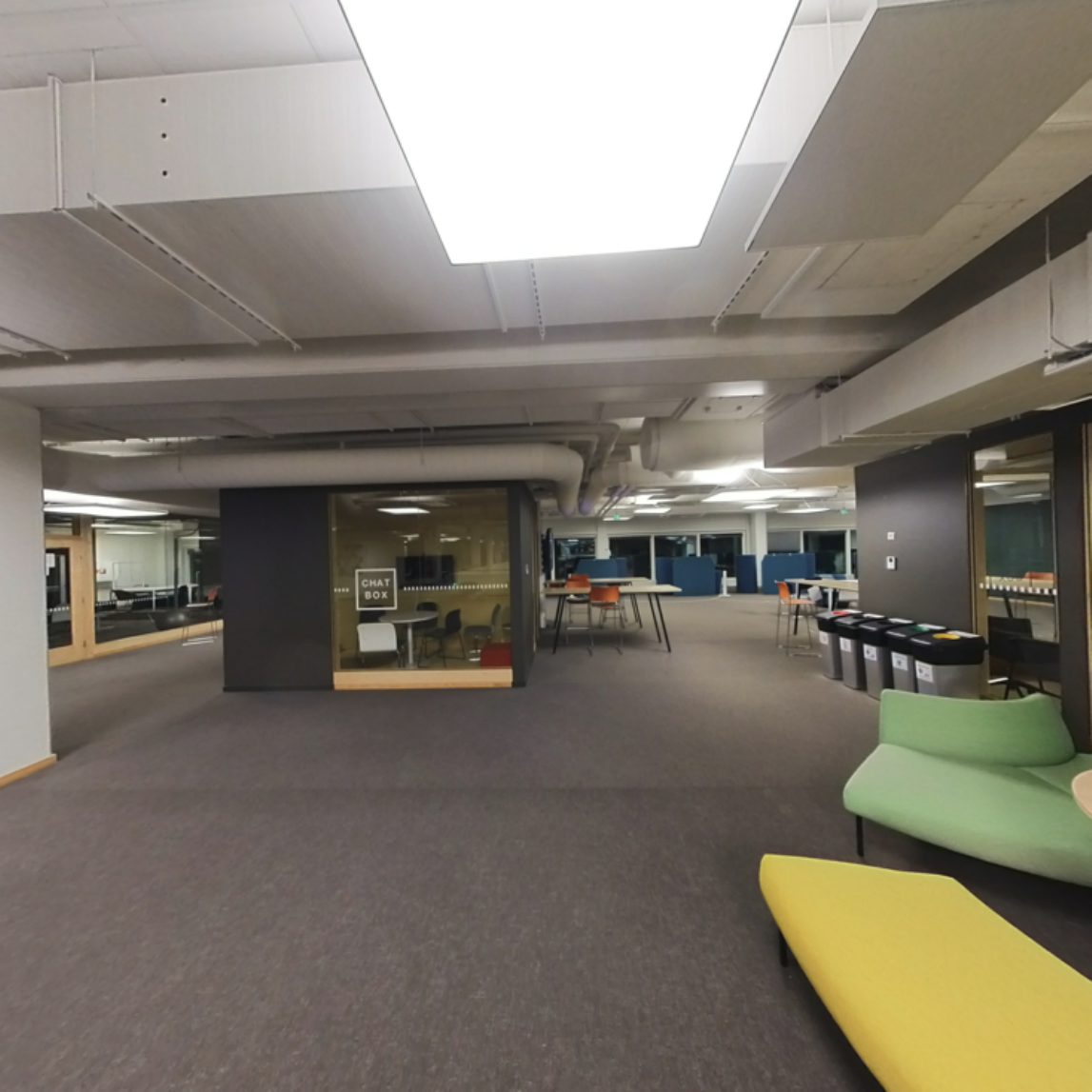}
        \end{minipage}
        \hfill
        \begin{minipage}{0.24\textwidth}
            \centering
            \textbf{Hall\_In} \\[1mm]
            \includegraphics[width=\textwidth]{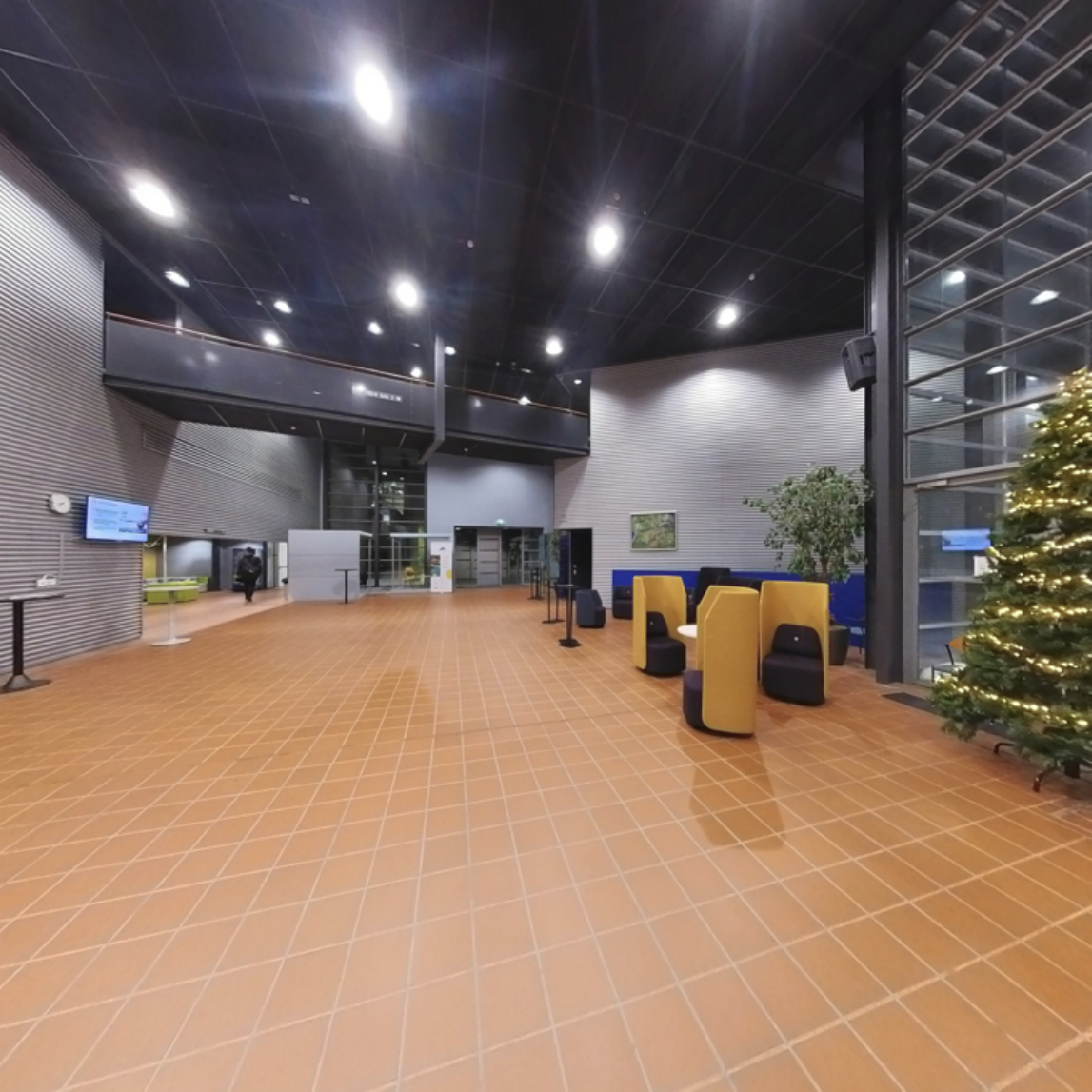}
        \end{minipage}

        \vspace{2mm} 

        \begin{minipage}{0.24\textwidth}
            \centering
            \includegraphics[width=\textwidth]{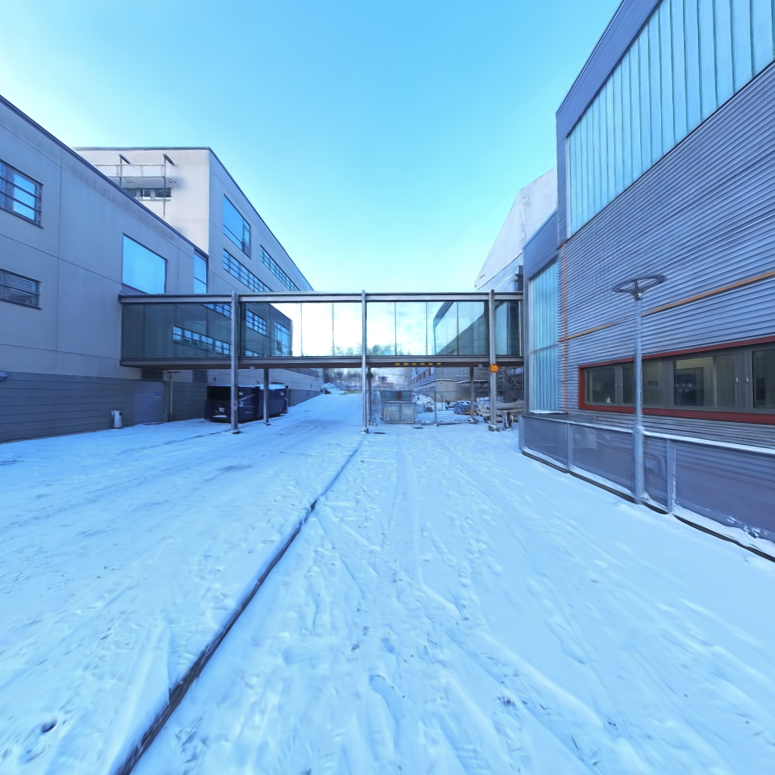}
        \end{minipage}
        \hfill
        \begin{minipage}{0.24\textwidth}
            \centering
            \includegraphics[width=\textwidth]{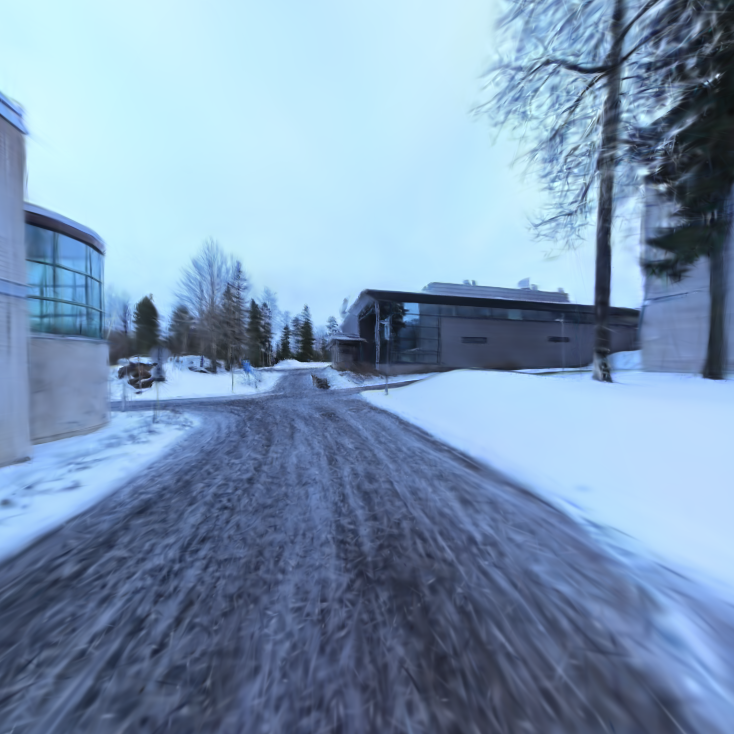}
        \end{minipage}
        \hfill
        \begin{minipage}{0.24\textwidth}
            \centering
            \includegraphics[width=\textwidth]{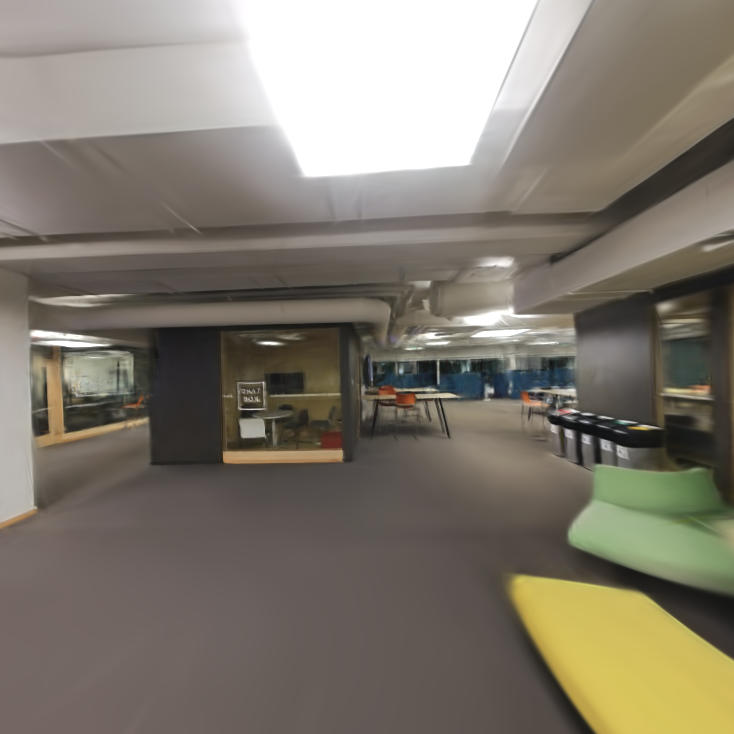}
        \end{minipage}
        \hfill
        \begin{minipage}{0.24\textwidth}
            \centering
            \includegraphics[width=\textwidth]{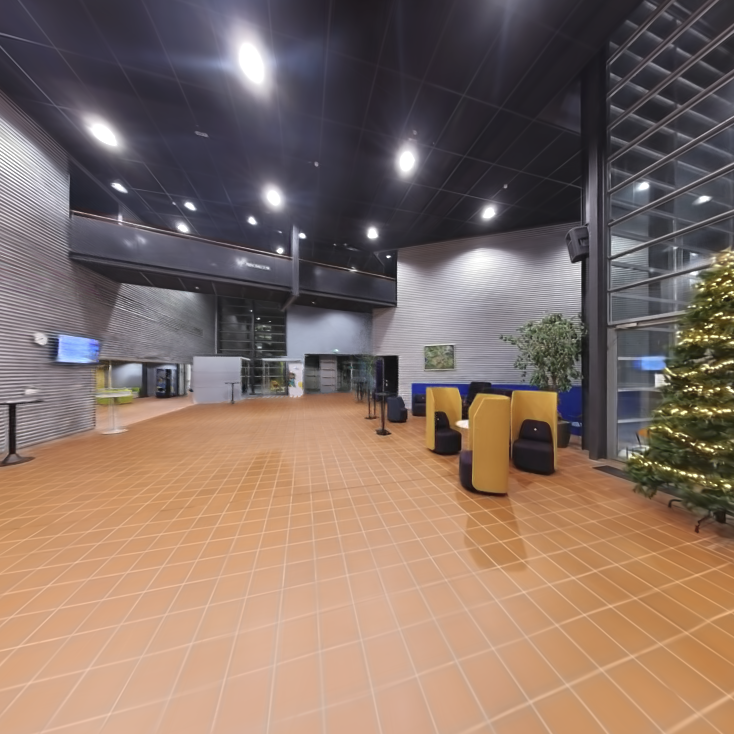}
        \end{minipage}
    \end{minipage}
\end{minipage}
}
\caption{\textbf{Compact comparison of Ground Truth vs.\ Gaussian Splatting renders for four representative scenes.}}
\label{fig:compare_gs}
\end{figure}

\cref{fig:compare_gs} presents example image-based rendering results obtained with the Gaussian splatting method for four scenes in our dataset. Each scene is illustrated by a single example to highlight the versatility of the data set across various environments. Example video renders of two of our scenes created from the Nerfacto model are also provided in the Supplementary Materials. Meanwhile, \cref{tab:metrics_10scenes} provides standard image quantitative metrics (PSNR, SSIM, and LPIPS\cite{zhang2018perceptual}) for all scenes in the dataset, averaged over the test images of each scene for both models. 

\begin{table}[htpb]
    \centering
    \small
    \caption{Quantitative Metrics (PSNR, SSIM, LPIPS) of 3DGS Method}
    \label{tab:metrics_10scenes}
    \begin{tabular}{|c|ccc|c|ccc|}
    \hline
    \multicolumn{4}{|c|}{\textbf{Indoor Scenes}} & \multicolumn{4}{c|}{\textbf{Outdoor Scenes}} \\
    \hline
    \textbf{Scene} & \textbf{PSNR} & \textbf{SSIM} & \textbf{LPIPS} & \textbf{Scene} & \textbf{PSNR} & \textbf{SSIM} & \textbf{LPIPS} \\
    \hline
    Upstairs\_In   & 23.33 & .8187 & .3693 & Bridge\_Out  & 27.58 & .8426 & .2544 \\
    Hall\_In     & 25.47 & .8354 & .1961 & Corridor\_Out    & 28.06 & .8507 & .2331 \\
    Building\_In & 26.28 & .8076 & .3017 & Building\_Out & 24.44 & .7525 & .3000 \\
    MeetingRoom\_In& 27.41 & .8628 & .2133 & Road\_Out  & 25.67 & .8109 & .2882 \\
    Kitchen\_In    & 27.15 & .8705 & .2200 & Night\_Out & 26.12 & .8328 & .3437 \\
    \hline
    \end{tabular}
\end{table}
\begin{table}[htpb]
    \centering
    \small
    \caption{Quantitative Metrics (PSNR, SSIM, LPIPS) of Nerfacto Model}
    \label{tab:metrics_nerfacto}
    \begin{tabular}{|c|ccc|c|ccc|}
    \hline
    \multicolumn{4}{|c|}{\textbf{Indoor Scenes}} & \multicolumn{4}{c|}{\textbf{Outdoor Scenes}} \\
    \hline
    \textbf{Scene} & \textbf{PSNR} & \textbf{SSIM} & \textbf{LPIPS} & \textbf{Scene} & \textbf{PSNR} & \textbf{SSIM} & \textbf{LPIPS} \\
    \hline
    Upstairs\_In   & 18.64 & .7768 & .5384 & Bridge\_Out  & 22.07 & .7799 & .3362 \\
    Hall\_In     & 21.30 & .7039 & .4590 & Corridor\_Out    & 18.91 & .5576 & .5630 \\
    Building\_In & 24.49 & .7936 & .4878 & Building\_Out & 20.31 & .4549 & .4081 \\
    MeetingRoom\_In& 23.90 & .8428 & .2622 & Road\_Out  & 19.15 & .6402 & .5086 \\
    Kitchen\_In    & 21.28 & .7922 & .3386 & Night\_Out & 18.19 & .6327 & .4979 \\
    \hline
    \end{tabular}
\end{table}

The baseline results from the vanilla Gaussian Splatting (3DGS) and Nerfacto-based scene reconstruction methods demonstrate the dataset's immediate applicability for novel view synthesis and 3D reconstruction tasks. As illustrated in \cref{fig:compare_gs}, the Gaussian Splatting method effectively handles varying lighting conditions and complex reflections, such as glare from glass surfaces and produces high-quality renders. Quantitative metrics in \cref{tab:metrics_10scenes} and \cref{tab:metrics_nerfacto}  further validate the robustness of these two models, with 3DGS providing better quantitative results than Nerfacto for our scenes.
\vspace{-3mm}
\subsection{Using dense LIDAR data for Gaussian scene initialization}
\begin{figure}[ht]
\centering
\small  
\setlength{\tabcolsep}{3pt} 
\begin{tabular}{c c c}
    \textbf{Ground Truth} & \textbf{COLMAP} & \textbf{COLMAP+LIDAR} \\
    
    \includegraphics[width=0.3\textwidth]{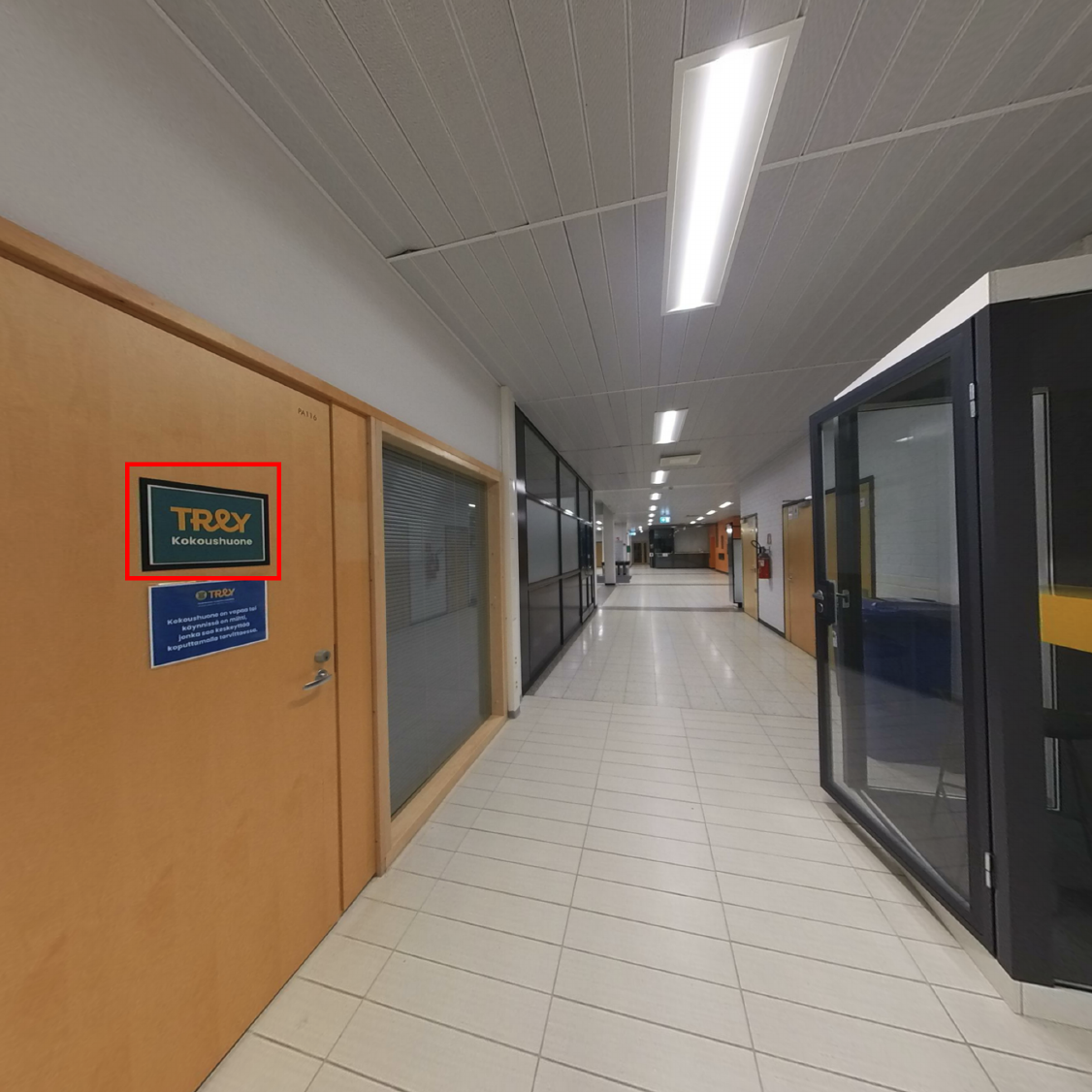} &
    \includegraphics[width=0.3\textwidth]{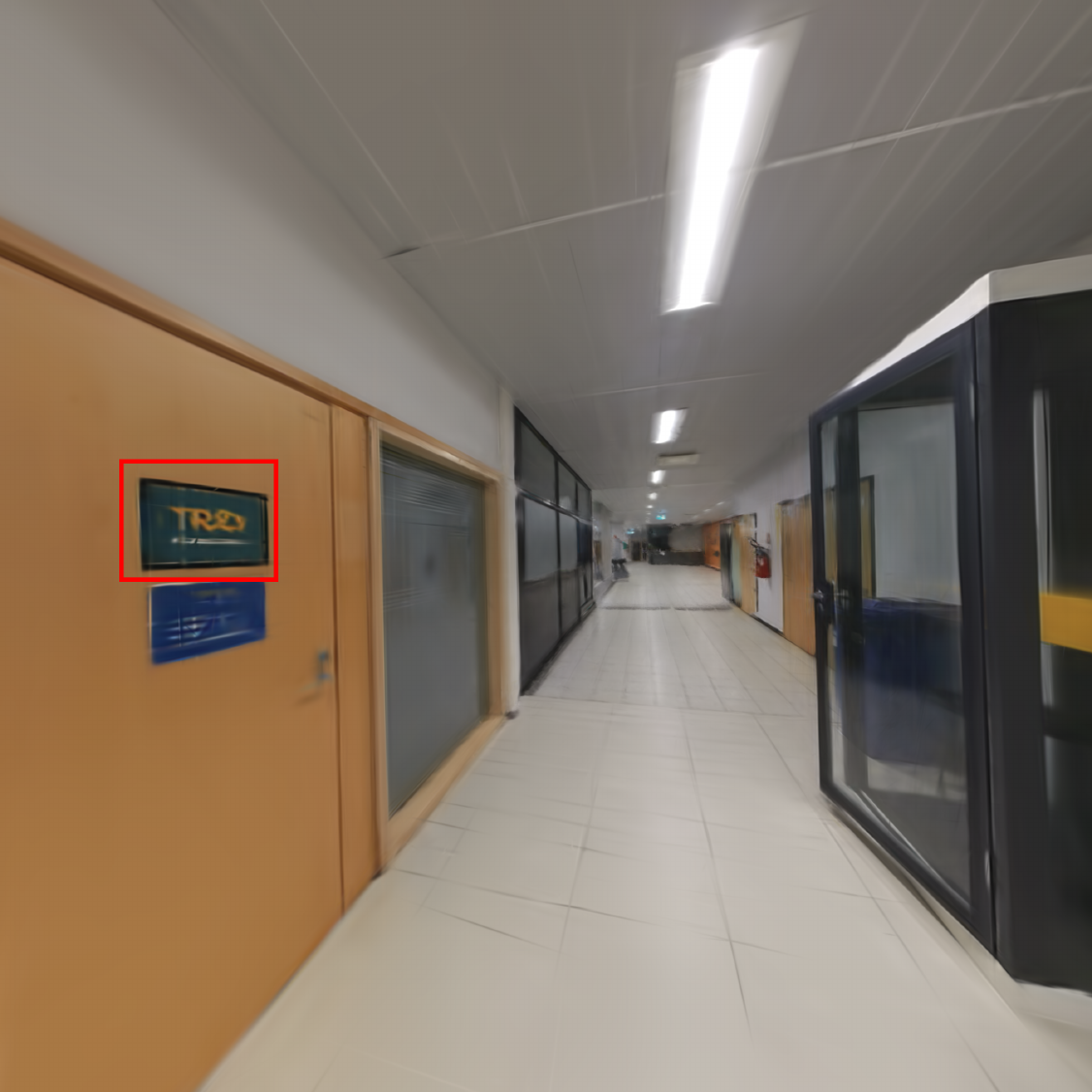} &
    \includegraphics[width=0.3\textwidth]{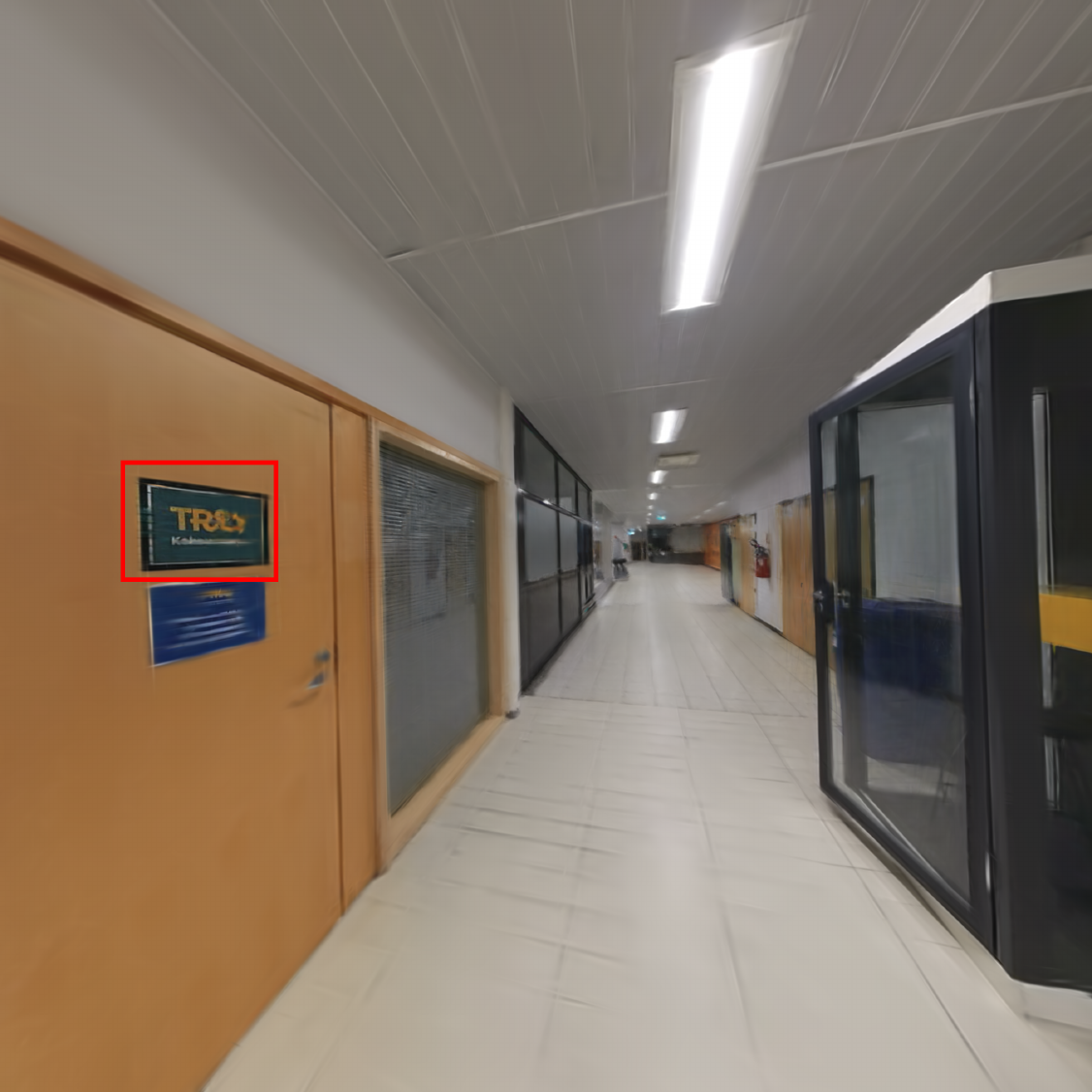} \\
    
    \includegraphics[width=0.3\textwidth]{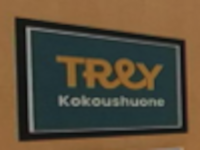} &
    \includegraphics[width=0.3\textwidth]{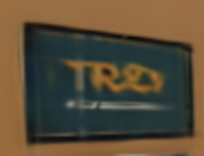} &
    \includegraphics[width=0.3\textwidth]{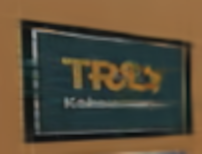} \\
\end{tabular}

\caption{
    \textbf{Comparison of an image rendering result for the Building\_In scene.} The ground truth image is compared against the rendering initialized with only COLMAP data versus the fused COLMAP+LIDAR point cloud.
}
\label{fig:Building}
\end{figure}
\vspace{-5mm}
\begin{table}[htbp]
\centering 
\caption{COLMAP vs. COLMAP+LIDAR Point Clouds for Gaussian Initialization on the Building\_In Scene (averaged over all test images)}
\label{tab:metrics_colmap_LIDAR}
\begin{tabular}{|c|c|c|c|c|}
\hline
\textbf{Gaussian Initialization} & \textbf{Number of Points (M)} & \textbf{PSNR} & \textbf{SSIM} & \textbf{LPIPS} \\
\hline
COLMAP         & $\sim$ 0.40 & 25.31 & 0.802 & 0.302 \\
COLMAP+LIDAR   & $\sim$ 2.6 & 26.24 & 0.811 & 0.269 \\
\hline
\end{tabular}
\end{table}
In the second experiment, we incorporate the dense point cloud we obtained from the Faro scanner to enrich the sparse point cloud provided by COLMAP. We begin by uniformly sub-sampling the Faro data to keep it computationally manageable, limiting the maximum available points in the dense point cloud to 2--3 times the points available in the sparse point cloud for easy and sufficiently accurate alignment, and then scale this sub-sampled cloud to match the scale of the COLMAP point cloud. After a rough manual alignment of rotation and translation, we register the sampled dense point cloud with Iterative Closest Point (ICP) \cite{Park_2017_ICCV}, yielding a fused point cloud. This LIDAR-aided point cloud serves as initialization for Gaussian splatting. Compared to the sparse point cloud, this fused point cloud enables the adaptive Gaussian training process to start from a better representative set of points. An example image-based rendering result of this densification process is shown from our ``Building\_In'' scene, in \cref{fig:Building}. For a true comparison, the test image set remains the same for both the sparse and fused approaches during rendering.

This experiment demonstrates that incorporating dense Faro LIDAR point clouds could improve both quantitative metrics and visual reconstruction quality, particularly in complex scenes. Beyond our experiment demonstrating a use case for the Faro scanner LIDAR data, there are many other potential uses for researchers, including but not limited to geometric accuracy evaluation, depth-based benchmarking, environment mapping and scene understanding.
\vspace{-3mm}
\section{Conclusion}
\vspace{-2mm}
We introduced a high-quality dataset to address limitations in existing resources for large-scale 3D scene reconstruction, novel view synthesis, and image-based rendering. Captured using an Insta360 camera with dual 200-degree fisheye lenses, the dataset provides comprehensive 360-degree coverage while compensating for heavy lens distortion through calibration and maintaining high scene detail. Complemented by dense ground truth point clouds from a Faro Focus 3D LiDAR scanner, it enables robust geometric evaluation and alignment benchmarking. The dataset presents unique challenges, which makes it well suited to 3D scene reconstruction under real-world complexities. By relying on still images, it avoids motion blur, maintains high detail, and provides a solid basis for advancing 3D reconstruction and novel view synthesis in complex environments.
\vspace{-2mm}
\section{Acknowledgements}
We acknowledge the financial support of the Intelligent Work Machines Doctoral Education Pilot Program (IWM VN/3137/2024-OKM-4) and funding from the Research Council of Finland (grants 352788, 353138, 362407, 362408, 339730, 353139, 362409) and the Finnish Center for Artificial Intelligence. We also acknowledge the Centre for Immersive Visual Technologies for the equipment used.  

%
%
%
\newpage
\bibliographystyle{splncs04}
\bibliography{name}
%




\end{document}